\documentclass[letterpaper]{article} 
\usepackage{aaai24}  
\usepackage{times}  
\usepackage{helvet}  
\usepackage{courier}  
\usepackage[hyphens]{url}  
\usepackage{graphicx} 
\usepackage{natbib}  
\usepackage{caption} 
\usepackage{algorithm}
\usepackage{algorithmic}

\usepackage{amsmath} 
\usepackage{bm}      
\usepackage{booktabs}
\usepackage{multirow}
\usepackage{xcolor}
\usepackage{placeins}

\usepackage{newfloat}
\usepackage{listings}
\DeclareCaptionStyle{ruled}{labelfont=normalfont,labelsep=colon,strut=off} 
\floatstyle{ruled}
\newfloat{listing}{tb}{lst}{}
\floatname{listing}{Listing}
\title{AMD: Anatomical Motion Diffusion with Interpretable Motion Decomposition and Fusion}
\author {
    Beibei Jing,
    Youjia Zhang,
    Zikai Song,
    Junqing Yu,
    Wei Yang \thanks{indicates the corresponding author}
}
\affiliations {
    Huazhong University of Science and Technology, Wuhan, China\\
    \{jingbeibei,youjiazhang, skyesong, yjqing, weiyangcs\}@hust.edu.cn
}

\usepackage{bibentry}

\usepackage{bibentry}

\let\oldtwocolumn\twocolumn
\renewcommand\twocolumn[1][]{%
    \oldtwocolumn[{#1}{
    
\begin{center}
\vspace{-1.5em}
\includegraphics[width=0.98\textwidth]{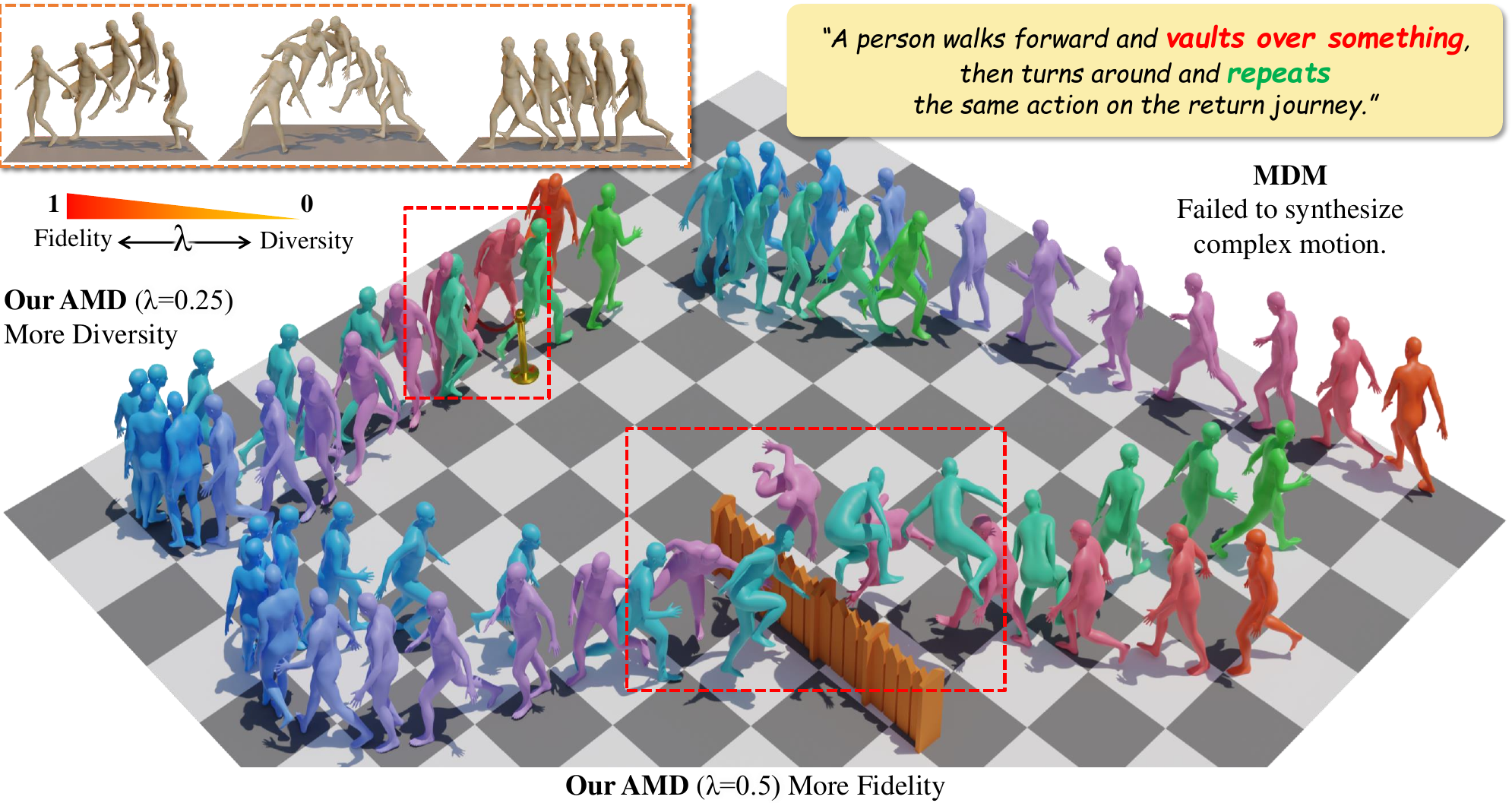}
    \captionof{figure}{
        Given a textual prompt, we employ a Large Language Model (LLM) to decompose it into anatomical scripts for targeted motion generation. Our method bypasses the intricacies of natural language encoding by converting comprehensive text into a singular vector. We introduce a fusion strategy to equitably condition both the comprehensive text (for fidelity) and the anatomical scripts (for diversity) during inverse diffusion. This technique effectively handles extended text inputs with intricate motions while maintaining a balance between fidelity and diversity.
    }
    \label{fig:teaser}
\end{center}
    }]
}

\begin{document}
\maketitle

\begin{abstract}
Generating realistic human motion sequences from text descriptions is a challenging task that requires capturing the rich expressiveness of both natural language and human motion. Recent advances in diffusion models have enabled significant progress in human motion synthesis. However, existing methods struggle to handle text inputs that describe complex or long motions. In this paper, we propose the Adaptable Motion Diffusion (AMD) model, which leverages a Large Language Model (LLM) to parse the input text into a sequence of concise and interpretable anatomical scripts that correspond to the target motion. This process exploits the LLM’s ability to provide anatomical guidance for complex motion synthesis. We then devise a two-branch fusion scheme that balances the influence of the input text and the anatomical scripts on the inverse diffusion process, which adaptively ensures the semantic fidelity and diversity of the synthesized motion. Our method can effectively handle texts with complex or long motion descriptions, where existing methods often fail. Experiments on datasets with relatively more complex motions, such as CLCD1 and CLCD2, demonstrate that our AMD  significantly outperforms existing state-of-the-art models.
\end{abstract}

\section{Introduction}

Human motion generation has witnessed remarkable advancements in recent years, largely driven by the emergence of sophisticated language models \citep{devlin2018bert,anil2023palm} and the innovation of diffusion generation techniques \citep{ho2020denoising,song2020denoising}. These breakthroughs have culminated in the development of motion generation models capable of crafting diverse and high-quality sequences, modulated by either textual \citep{zhang2022motiondiffuse,ren2023diffusion,shafir2023human} or other control inputs \citep{gong2023tm2d,zhao2023modiff}. With applications already spreading into the gaming industry \cite{zhang2022wanderings}, these models present a promising frontier for both novices and professionals in character animation.

Traditional methods employ variational autoencoder (VAE) structures to encode conditions and 3D human motions separately, subsequently learning the joint distribution from text-motion paired datasets \citep{petrovich2022temos,lee2023multiact}. Recent strategies harness advances in diffusion models \citep{ho2020denoising,song2020denoising,tevet2022human}, treating motion synthesis as an inverse diffusion process \citep{chen2023executing,ao2023gesturediffuclip,yuan2022physdiff}. Additionally, some research transforms human motion features into a codebook, iteratively predicting the motion index and reconstructing the entire motion \citep{guo2022tm2t,zhang2023t2m,lucas2022posegpt}.

Despite these strides, existing methods frequently oversimplify the complexity of language, utilizing a text encoder to transform text prompts into a fixed-length vector, regardless of text length or rarity. Such an approach demands large-scale datasets to encompass the full mapping spectrum and is prone to failure with unseen text related to complex or extended sequential actions. For example, the term `scorpion' refers to an expert dance move described by the specific motions `flex and grasp foot, swing leg around, lock out arm'. Clearly, synthesizing this rare `scorpion' motion is intractable without adequate training, yet the detailed description offers abundant instructions about its execution in a common language. This issue is analogous to lengthy motion texts, which could be dissected into concise anatomical scripts.

Inspired by this observation, we introduce the innovative Adaptable Motion Diffusion (AMD) model. It leverages the capabilities of a Large Language Model (LLM), specifically a finetuned ChatGPT-3.5 in our implementation, to parse input text into sequences of concise, interpretable anatomical scripts, synergized with the generative prowess of diffusion models. Our approach capitalizes on the LLM's capacity to elucidate complex motion synthesis via anatomical explanations. We construct a two-branch diffusion process wherein one branch uses the source text as a condition, and the other depends on the dissected anatomical scripts, and corresponding reference motions retrieved from a motion database using the anatomical scripts.The database consists of text descriptions, decomposed text, and corresponding action sequences in our implementation. The inverse diffusion processes in both branches yield two motion sequences that align with the source text and anatomical scripts, respectively. We then blend these motions through a meticulously designed fusion block, generating the final motion sequence.

Extensive experiments demonstrate that our approach rivals existing state-of-the-art models and particularly excels with lengthy and intricate text. Furthermore, our model exhibits the ability to generate human actions under various scenarios—conditioned, partially conditioned, or unconditioned—thereby augmenting its robustness. It also encompasses a partial action editing feature akin to the Motion Diffusion Model (MDM), facilitating the alteration of primary human action sequences through detailed descriptions. This functionality fosters the generation of more complex and nuanced actions.

\begin{figure*}[t]
\centering
\includegraphics[width=0.95\textwidth]{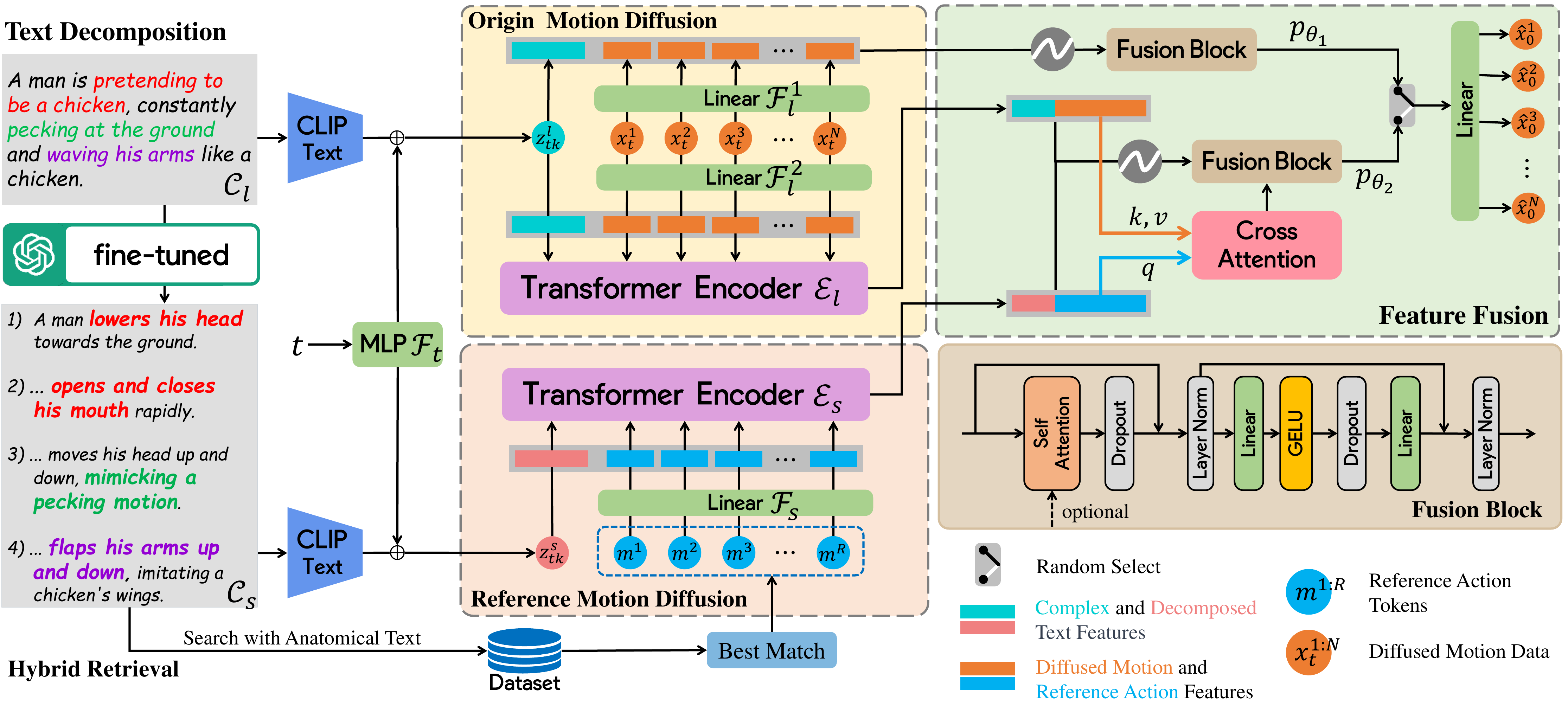} 
  \caption{Overview of our method architecture:  Given a motion text prompt ${c_l}$, we decompose it into a set of anatomical scripts ${c_s}$ using a fine-tuned ChatGPT model. Then, we use a MLP ${F_t}$ to project ${c_l}$ , ${c_s}$ and $t$ into tokens $z^{l}_{tk}$ and  $z^{s}_{tk}$, respectively. Next, we exploit the guidance from anatomical scripts ${c_s}$ by searching for reference motions ${m}^{1: R}$ from a database using the anatomical similarity. In the Reference Motion Diffusion (RMD) module, we alter the dimensions of ${m}^{1: R}$ with ${F_s}$ and send it along with $z^{s}_{tk}$ into ${E_s}$ to obtain decomposed text features and reference action features; similarly, in the Origin Motion Diffusion (OMD) module, we use $F_{l}^{2}$ to alter the dimensions of $x_{t}^{1:N}$ and send it along with $z^{l}_{tk}$ into ${E_l}$ to obtain canonical text features and diffused motion features. In the Feature Fusion (FF) module, we send the decomposed text features, reference action features, complex text features, and diffused motion features, all processed through positional embedding, along with the diffused action features and reference action features treated through cross-attention mechanism into the Fusion Block, yielding the output $p_{\theta_2}$; concurrently, we use $F_{l}^{1}$ to alter the dimensions of $x_{t}^{1:N}$ and send it along with $z^{l}_{tk}$, post positional encoding, directly into Fusion Block, yielding $p_{\theta_1}$. Finally, we adaptively adjust the output frequencies of $p_{\theta_1}$ and $p_{\theta_2}$ using \(\lambda\) (see Func. \ref{Fusion}).The Fusion Block, designed on top of the Transformer Encoder Layer, incorporates residual connections.}
\label{Overview of the architecture}
\end{figure*}

\section{Related Work}
\textbf{Human Motion Generation.} 
Existing human motion generation research can be primarily categorized into two groups: 1) unconstrained generation, with the goal of producing motion sequences free from explicit conditioning factors~\citep{mukai2005geostatistical,rose1998verbs}; and 2) conditional synthesis, which focuses on achieving controllability such as Language2Pose \cite{ahuja2019language2pose},Text2Gesture \cite{bhattacharya2021text2gestures} and Dance2Music \cite{lee2019dancing}.
Conditional motion generation is of particular interest to many researchers. The utilization of sequence-to-sequence RNN models has been explored by \cite{lin2018generating}. And \cite{ghosh2021synthesis} takes into account the structure of the human skeleton and puts forward a hierarchical two-stream approach for pose generation. Besides,\cite{guo2022generating} proposes a two-step approach involving text2length sampling and text2motion generation. \cite{lee2023multiact} and TEMOS \cite{petrovich2022temos} suggest using a VAE \cite{kingma2013auto} to map a text prompt into a normal distribution in latent space.
The diffusion models \cite{ho2020denoising}, which gradually denoise a sample from the data distribution, have been applied to human motion synthesis by recent methods PhysDiff \cite{yuan2022physdiff}, MDM \cite{tevet2022human}, MoFusion \cite{dabral2023mofusion} and ReMoDiffuse \cite{zhang2023remodiffuse}. MDM and PhysDiff have enriched conditional generation tasks, with MDM highlighting geometric losses for enhanced results, and PhysDiff incorporating physical constraints for realistic motion. An innovative task has been proposed for generating 3D dance motion using both textual and musical modalities \cite{gong2023tm2d}. Moreover, the MoFusion framework leverages denoising diffusion for high-quality conditional human motion synthesis, while ReMoDiffuse employs a denoising process for action generation via a retrieval mechanism, illustrating the versatility and potential of diffusion models in this field.
In this paper, we leverage retrieved approximate actions and textual descriptions segmented through the ChatGPT-3.5 API, inspired by ReMoDiffuse \cite{zhang2023remodiffuse}. We then map these textual descriptions to corresponding human body motions using an inverse diffusion process. Experiments on complex text test sets CLCD1 and CLCD2 show our model significantly outperforms existing models in handling complex text.

\textbf{Diffusion Generative Models}
The diffusion generative models, a promising direction in image synthesis, have also been adapted for human motion synthesis, sparking a variety of innovative methodologies. Diffusion generative models, first introduced by DDPM \cite{ho2020denoising} and further refined by DDIM \cite{song2020denoising}, have gained considerable traction in the field of image synthesis, as demonstrated by works like Imagen \cite{saharia2022photorealistic} and Stable Diffusion \cite{rombach2022high}.The core principle of diffusion generative models involves a sophisticated process of adding noise to data, diffusing it into a simpler form, and then reversing this process to create new data similar to the original.

\textbf{Human Motion Dataset and Large Language Models}
The importance of motion data is paramount in motion synthesis tasks. Tools like the KIT Motion-Language \cite{mandery2015kit} and HumanML3D \cite{guo2022generating} have become essential in the text-to-motion task. KIT Motion-Language offers sequence-level descriptions for various motions extracted from another dataset \cite{mandery2015kit}, and HumanML3D augments certain motions from AMASS \cite{mahmood2019amass}  and HumanAct12 \cite{guo2020action2motion} with extensive textual annotations (encompassing 14,616 motions annotated with 44,970 textual descriptions). These datasets have been utilized in leading-edge text-driven motion generation research, as evidenced by works \citep{chen2023executing,kim2023flame}.

ChatGPT-3.5 developed by OpenAI, is renowned for creating coherent, natural text. Utilizing the Transformer architecture and a vast, varied dataset, it's adept at interpreting human language in diverse contexts. Unlike models tailored for specific tasks, ChatGPT's design emphasizes broad applicability and continuous refinement across language domains.

\section{Anatomical Motion Diffusion}

We commence with the motion text anatomical decomposition procedure, followed by a discussion on the two branch diffusion and fusion, and finally the model's training and inference process.Our model is illustrated in Figure \ref{Overview of the architecture}.

\subsection{Anatomical Motion Text Decomposition}
Human language is highly complex and multi-dimensional representation. In the task of text-to-human motion synthesis, the correct representation and mapping of textual descriptions to features are crucial for the synthesis of motion. However, previous work usually uses a single text encoder to map the text into a fixed length vector, no matter how long or complex the text is, and overlooks the importance of text processing. We propose to decompose the text prompt into a set of anatomical scripts to bypass the obstacles imposed by language complexity. 

Specifically, we leverage the power of a Large Language Model, i.e., a fine-tuned ChatGPT-3.5 model, for decomposing complex action text descriptions into a superposition of several simple anatomical text scripts (each description containing only a simple action), as illustrated in the text decomposition process on the left side of Figure \ref{Overview of the architecture}.

For finetuning the ChatGPT-3.5 model, we proceed the following step by step:
\begin{itemize}
    \item First, we used the ChatGPT-3.5 API to break down the text in the dataset into multiple texts containing simple actions, such as ``a person raising a hand'' and ``walking''. The specific prompt is ``Please decompose the following action into simple actions:Sentence ''.
    \item Then, to align with the original text description and be concise, we selected the simple anatomical scripts that must be included in that text description.
    \item Afterwards, we formed text pairs \((c_l, c_s)\) consisting of the original text and the simple action descriptions that must be included, thus constituting all the data used for fine-tuning.
\end{itemize}
During the testing process, the model is able to decompose complex action text into anatomical scripts containing simple actions per script. We use the Sentence-BERT model all-MiniLM-L6-v2 to compute the semantic embeddings of texts and anatomical scripts.

\subsection{Motion Diffusion and Fusion}
We leverage the advanced diffusion models for synthesizing motion sequences from text. We carefully design two diffusion models to synthesize motion from the original comprehensive text (for fidelity) and the decomposed anatomical scripts (for diversity), and construct a fusion module to adaptively balance the two motions. For decomposed anatomical scripts, we search for similar anatomical texts in a database, and fetch the corresponding motion as a reference of the diffusion.

\subsubsection{Canonical Diffusion}
The genesis of the Denoising Diffusion Probabilistic Model (DDPM) diffusion process is rooted in a simplistic prior, generally a Gaussian distribution, that successively transforms into a complex data distribution through the additive injection of noise at individual temporal steps. Mathematically, this process can be characterized as a series of conditional probability distributions, with each stage simulated by a denoising process parameterized by a neural network.
Specifically, the Canonical Diffusion is characterised by a subsequent conditional distribution:

\begin{equation}
q(x_{t}^{1:N}|x_{t-1}^{1:N}) = \mathcal{N}(x_{t}^{1:N}; \sqrt{1 - \beta_t}x_{t-1}^{1:N}, \beta_t\textbf{\textit{I}}
).
\end{equation}

The reverse of the diffusion process is considered by the model, using the following parameterization:
\begin{equation}
p_{\theta_1}(x_{t-1}^{1:N}|x_{t}^{1:N}, c) = \mathcal{N}(x_{t-1}^{1:N}; \boldsymbol{\mu}_{\theta}(x_{t}^{1:N}, c, t),\boldsymbol{\Sigma}_{\theta}(x_{t}^{1:N}, c, t)) \label{Canonical Diffusion}
\end{equation}

Where \( x_{0}^{1:N} \), \( x_{t}^{1:N} \) and \( x_{T}^{1:N} \) represent  clear motion frames, diffused motion data (the state at time step \( t \))  and the fully-diffused Gaussian noise, respectively, \( t = 1, 2, \ldots, T \). $c$ is an optional conditional variable, in this case referring to textual $c_l$. \(N\) represents the number of frames in the motion sequence. \(\beta_t \in (0, 1)\) is a non-negative diffusion coefficient that controls the difference between the current state and the previous state.

\subsubsection{Reference Diffusion}
In reference diffusion, to enable the model to manage intricate text descriptions, we incorporate both the decomposed short text and reference action information as conditions. These are then added to the inversion of the diffusion process. As a result, Equation~\ref{Canonical Diffusion}  is reformulated as:
\begin{align}
p_{\theta_2}(x_{t-1}^{1:N}|x_{t}^{1:N},c) &= \mathcal{N}(x_{t-1}^{1:N}; \boldsymbol{\mu}_{\theta}(x_{t}^{1:N},c, m^{1: R}, t), \nonumber \\
&\quad \boldsymbol{\Sigma}_{\theta}(x_{t}^{1:N}, c,t)).\label{Reference Diffusion}
\end{align}
Where \({c}\) denote the textual condition $c_l$, the decomposed textual condition $c_s$ , and \({m}^{1: R}\) (represents the reference action tokens), with \(R\) as the frame count.

\subsubsection{Fusion}
The purpose of the fusion module is to adaptively balance the motions generated by the caninocal diffusion and the reference diffusion. Figure~\ref{Overview of the architecture} illustrates two complementary methodologies for motion generation, integrating the diffusion model. Our objective can be expressed through the equations:

\begin{equation}
f(p_{\theta},\lambda) = 
\begin{cases} 
1-\lambda   &\mathrm{if} \  p_{\theta_1}  \\
\lambda   &\mathrm{if}\  p_{\theta_2} \label{Fusion}
\end{cases}
\end{equation}

Where \(f(\cdot)\) is Bernoulli distribution,  \(\lambda\) is a probability value, used to control the frequency of the outputs from two different pipelines. Specifically, \(p_{\theta_ 1}\) and \(p_{\theta_2}\) represent the output  of two different pipelines, respectively. By adjusting \(\lambda\), we can precisely control the frequency of these two outputs.

\subsection{Training and Inference}
\subsubsection{Model Training:}
We adopt a novel approach in the training phase, using elements of a classifier-free technique. Specifically, for unconditional generation, we randomly mask 10\% of the textual conditions and related action information. Additionally, we approximate the probability distribution  \(x_0^{1: N}\) using a mixture of elements, and we random select $p_{\theta_1}$ and $p_{\theta_2}$. The primary goal of the training process is to minimize the mean square error between the predicted initial sequence and the actual ground truth, thereby enhancing the model's accuracy.During the retrieval process, we obtain the action sequences closest to the input by calculating the semantic similarity between the decomposed texts of the input complex text and those in the database.

\begin{figure*}[t]
\centering
\includegraphics[width=0.95\textwidth]{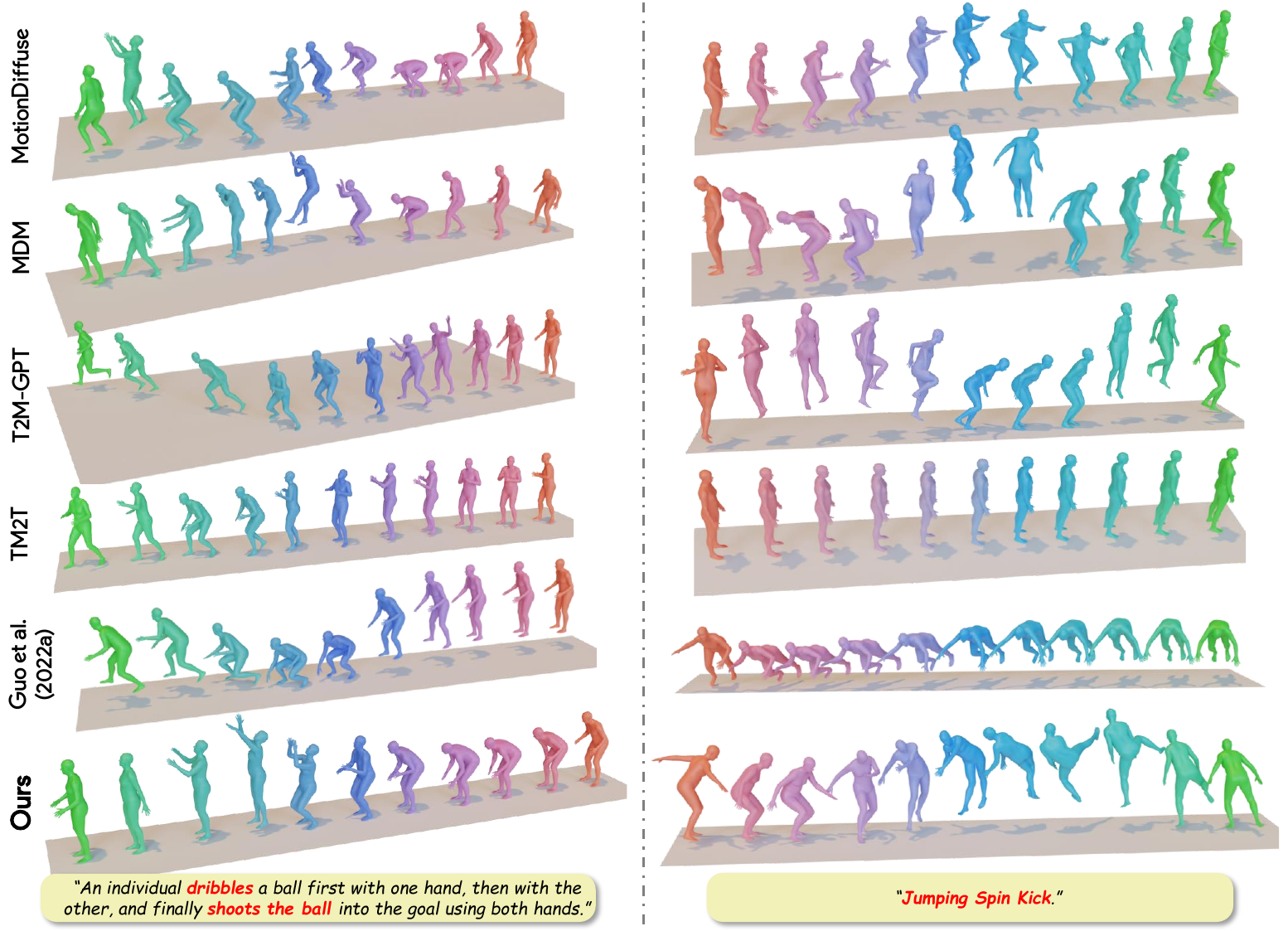} 
\caption{AMD shows strong ability in handling protracted text prompt (left) and semantic rich text (right). The motion sequences progress from red to green.}
\label{Text to action synthesis_dribbles a ball}
\end{figure*}

\subsubsection{Model Inference:}
 In the reverse stage, we first obtain the predicted sequence \( \widehat{{x_0}}^{1: N} = G_i\left( {{x_t}^{1: N},c,t} \right) \) by feeding the AMD model with Gaussian noise \({x_T}^{1: N}\) and an optional condition, corresponds to Equation \ref{Fusion}. Then, we add noise to \(x_{t-1}\) and repeat the process until we reach \(x_0^{1: N}\). Furthermore, our model can generate movements without conditions or with partial conditioning.When sampling \(G\), we can use s to find a balance between diversity and fidelity by interpolating or extrapolating the two variants.

\begin{align}
A_s(x_t^{1: N}, c_l, t)) & = G_A(x_t^{1: N}, \emptyset, t) \notag\\
                         & + s \cdot ( G_A(x_t^{1: N}, c_l, t) \notag\\
                         & - G_A(x_t^{1: N}, \emptyset, t) ) \label{H_s}
\end{align}

In \(A_s\), \(G_A\) acquires knowledge of both the conditioned and unconditioned distributions. It accomplishes this by assigning \(c_l = \emptyset\) to 10\% of the samples at random, which allows \(G_A(x_t^{1: N}, \emptyset, t)\)  to serve as an approximation of \(p_{\theta}(x_0^{1: N})\). 

\begin{align}
B_s(x_t^{1: N}, c_l,c_s,m^{1: R}, t) & = G_B(x_t^{1: N}, \emptyset,c_s,m^{1: R}, t) \notag \\
                                     & + s \cdot ( G_B(x_t^{1: N}, c_l,c_s,m^{1: R}, t) \notag \\
                                     & - G_B(x_t^{1: N}, \emptyset,c_s,m^{1: R}, t) ) \label{J_s}
\end{align}

Similarly, by adding conditions, we can obtain  $B_s(x_t^{1: N}, c_l, c_s, m^{1: R},t)$.In \(B_s\), in order for the model to be adequately trained on simple text, this is achieved by randomly assigning \(c_l=\emptyset\) to 10\% of the samples, which allows \(G_B\) to serve as an approximation of \(p_{\theta}(x_0^{1: N})\).

\begin{equation}
f(G_s,\lambda) = 
\begin{cases} 
1-\lambda   &\mathrm{if} \ A_s(x_t^{1: N}, c_l, t)) \\
\lambda   &\mathrm{if}\ B_s(x_t^{1: N}, c_l, c_s, m^{1: R},t) 
\end{cases}
\end{equation}

In \(G_s\), we adjust the output frequencies of \(B_s\) and \(A_s\) by modifying the parameter \(\lambda \). This fine-tuning enables the model to strike a balance between Diversity and Fidelity, allowing for a trade-off that enhances performance. 

\begin{table*}[t]
\centering
\scalebox{0.9}{
\begin{tabular}{llccccc}
\toprule
Dataset & Method & R Precision (top 3)$\uparrow$ & FID$\downarrow$ & Multimodal Dist$\downarrow$ & Diversity$\rightarrow$ & Multimodality$\uparrow$ \\
\midrule
\multirow{8}{*}{\centering SLCD1} 
& Real & $0.731^{\pm.006}$ & $0.036^{\pm.004}$ & $3.243^{\pm.007}$ & $8.139^{\pm.036}$ & - \\
\midrule
& Guo et al. (2022a) & $0.646^{\pm.010}$ & $2.257^{\pm.127}$ & \textcolor{blue}{\textbf{$3.535^{\pm.026}$}} & \textcolor{blue}{\textbf{$7.998^{\pm.083}$}} & $2.259^{\pm.166}$ \\
& MDM & $0.569^{\pm.009}$ & \textcolor{blue}{\textbf{$0.828^{\pm.061}$}}& $4.927^{\pm.059}$ & $8.669^{\pm.083}$ &\textcolor{blue}{\textbf{$2.272^{\pm1.005}$}} \\
& TM2T & $0.651^{\pm.009}$ & $1.701^{\pm.078}$ & $3.614^{\pm.031}$ & $7.555^{\pm.063}$ & \textcolor{red}{\textbf{$2.477^{\pm.135}$ }}\\
& T2M-GPT & $0.659^{\pm.010}$ & $0.603^{\pm.073}$ & $3.566^{\pm.039}$ & $8.725^{\pm.088}$ & $1.809^{\pm.099}$ \\
& MotionDiffuse & \textcolor{red}{\textbf{$0.693^{\pm.011}$}} & $0.830^{\pm.043}$ &\textcolor{red}{\textbf{$3.338^{\pm.040}$}}  & $8.378^{\pm.088}$ & $1.619^{\pm.068}$ \\
\midrule
& Ours & \textcolor{blue}{\textbf{$0.673^{\pm.007}$}} & \textcolor{red}{\textbf{$0.116^{\pm.011}$}} & $4.296^{\pm.012}$ & \textcolor{red}{\textbf{$8.120^{\pm.084}$}} & $1.399^{\pm.077}$ \\
\cline{1-7}
\multirow{8}{*}{\centering SLCD2} 
& Real & $0.774^{\pm.003}$ & $0.004^{\pm.000}$ & $3.083^{\pm.001}$ & $8.666^{\pm.063}$ & - \\
\midrule
& Guo et al. (2022a) & $0.695^{\pm.004}$ & $1.547^{\pm.046}$ & $3.471^{\pm.013}$ & \textcolor{blue}{\textbf{$8.554^{\pm.077}$}} & $2.105^{\pm.091}$ \\
& MDM & $0.592^{\pm.006}$ & $0.540^{\pm.042}$ & $5.074^{\pm.029}$ & $8.865^{\pm.098}$ & \textcolor{red}{\textbf{$2.879^{\pm.092}$}} \\
& TM2T & $0.710^{\pm.004}$ & $1.359^{\pm.035}$ & $3.474^{\pm.014}$ & $7.970^{\pm.069}$ & \textcolor{blue}{\textbf{$2.456^{\pm.175}$}} \\
& T2M-GPT & \textcolor{blue}{\textbf{$0.722^{\pm.005}$}} & \textcolor{blue}{\textbf{$0.298^{\pm.011}$}} & \textcolor{blue}{\textbf{$3.381^{\pm.013}$}} &$9.112^{\pm.081}$ & $1.968^{\pm.065}$ \\
& MotionDiffuse & \textcolor{red}{\textbf{$0.742^{\pm.004}$}} & $0.804^{\pm.026}$ & \textcolor{red}{\textbf{$3.224^{\pm.014}$}} & $8.889^{\pm.104}$ & $1.433^{\pm.103}$ \\
\midrule
& Ours & $0.667^{\pm.005}$ & \textcolor{red}{\textbf{$0.146^{\pm.014}$}} & $4.619^{\pm.018}$ & \textcolor{red}{\textbf{$8.586^{\pm.083}$}} & $1.384^{\pm.066}$ \\
\bottomrule
\end{tabular}}
\caption{Quantitative results on the SLCD1 and SLCD2 test sets: All methods use the real motion length from the ground truth. $\rightarrow$ means results are better if the metric is closer to the real distribution. We ran all the evaluations 20 times (except MultiModality, which ran 5 times), and $\pm$ indicates the 95\% confidence interval. Red indicates the best result, while Blue indicates the second-best result.}
\label{tab:performance_SLCD1_2}
\end{table*}

\begin{table*}[ht]
\centering
\setlength\tabcolsep{1pt}
\scalebox{0.93}{
\begin{tabular}{l|ccccc|ccccc}
\hline
\multicolumn{1}{c|}{Method} & \multicolumn{5}{c|}{HumanML3D} & \multicolumn{5}{c}{KIT} \\
\hline
& R-Precision $\uparrow$ & FID $\downarrow$ & MM Dist $\downarrow$ & Diversity $\to$ & MMD $\updownarrow$ & R-Precision $\uparrow$ & FID $\downarrow$ & MM Dist $\downarrow$ & Diversity $\to$ & MMD $\updownarrow$ \\ 
\hline
Real & $0.797^{\pm.002}$ & $0.002^{\pm.000}$ & $2.974^{\pm.008}$ & $9.503^{\pm.065}$ & - & $0.779^{\pm.006}$ & $0.031^{\pm.004}$ & $2.788^{\pm.012}$ & $11.08^{\pm.097}$ & - \\ 
\hline
Guo et al.& $0.740^{\pm.003}$ & $1.067^{\pm.002}$ & $3.340^{\pm.008}$ & $9.188^{\pm.002}$ & $2.090^{\pm.083}$ & $0.693^{\pm.007}$ & $2.770^{\pm.109}$ & $3.401^{\pm.008}$ & $10.91^{\pm.119}$ & $2.077^{\pm.274}$ \\
MDM & $0.611^{\pm.007}$ & $0.544^{\pm.044}$ & $5.566^{\pm.027}$ & $9.559^{\pm.086}$ & \textcolor{blue}{\textbf{$2.799^{\pm.072}$}} & $0.396^{\pm.004}$ & $0.497^{\pm.021}$ & $9.191^{\pm.022}$ & $10.847^{\pm.109}$ & $1.907^{\pm.214}$ \\
TM2T & $0.729^{\pm.002}$ & $1.501^{\pm.017}$ & $3.467^{\pm.011}$ & $8.589^{\pm.076}$ & $2.424^{\pm.093}$ & $0.587^{\pm.005}$ & $3.599^{\pm.153}$ & $4.591^{\pm.026}$ & $9.473^{\pm.117}$ & \textcolor{blue}{\textbf{$3.292^{\pm.081}$}} \\
MMA & $0.676^{\pm.002}$ & $0.774^{\pm.007}$ & - & $8.23^{\pm.064}$ & - & - & - & - & - & - \\
MDF & \textcolor{blue}{\textbf{$0.782^{\pm.001}$}} & $0.630^{\pm.001}$ & \textcolor{blue}{\textbf{$3.113^{\pm.001}$}} & $9.410^{\pm.049}$ & $1.553^{\pm.042}$ & \textcolor{red}{\textbf{$0.739^{\pm.004}$}} & $1.954^{\pm.062}$ & \textcolor{blue}{\textbf{$2.958^{\pm.005}$}} & \textcolor{red}{\textbf{$11.10^{\pm.143}$}} & $0.730^{\pm.013}$ \\
ReMDF & \textcolor{red}{\textbf{$0.795^{\pm.004}$}} & \textcolor{red}{\textbf{$0.103^{\pm.004}$}} & \textcolor{red}{\textbf{$2.974^{\pm.016}$}} & $9.018^{\pm.075}$ & $1.795^{\pm.043}$ & $0.765^{\pm.055}$ & \textcolor{red}{\textbf{$0.155^{\pm.006}$}} & \textcolor{red}{\textbf{$2.814^{\pm.012}$}} & $10.80^{\pm.105}$ & $1.239^{\pm.028}$ \\
DiffKFC & $0.681^{\pm.005}$ & $0.148^{\pm.029}$ & $4.988^{\pm.022}$ & \textcolor{blue}{\textbf{$9.467^{\pm.087}$}} & \textcolor{red}{\textbf{$0.288^{\pm.021}$}} & $0.414^{\pm.006}$ & \textcolor{blue}{\textbf{$0.180^{\pm.028}$}} & $8.908^{\pm.012}$ & $10.97^{\pm.112}$ & \textcolor{red}{\textbf{$0.196^{\pm.052}$}} \\
T2M-GPT & $0.775^{\pm.002}$ & \textcolor{blue}{\textbf{$0.116^{\pm.004}$}} & $3.118^{\pm.011}$ & $9.761^{\pm.081}$ & $1.856^{\pm.011}$ & \textcolor{blue}{\textbf{$0.737^{\pm.006}$}} & $0.717^{\pm.041}$ & $3.053^{\pm.026}$ & $10.862^{\pm.094}$ & $1.912^{\pm.036}$ \\
\hline
Ours & $0.657^{\pm.006}$ & $0.204^{\pm.031}$ & $5.282^{\pm.032}$ & \textcolor{red}{\textbf{$9.476^{\pm.077}$}} & $1.356^{\pm.089}$ & $0.401^{\pm.005}$ & $0.233^{\pm.068}$ & $9.165^{\pm.032}$ & \textcolor{blue}{\textbf{$10.971^{\pm.126}$}} & $1.600^{\pm.174}$ \\ 
\hline
\end{tabular}}
\caption{Quantitative comparison of AMD and baselines on  HumanML3D and KIT datasets.(Guo et al.$\to$Guo et al. (2022a),MDF$\to$MotionDiffuse,ReMDF$\to$ReMoDiffuse,MMD$\to$Multimodality)}
\label{combined-dataset}
\end{table*}

\section{Experiments}

\subsection{Text-to-Motion}
\textbf{Datasets:}
In addition to evaluating KIT-ML and HumanML3D (as in \cite{guo2022generating}), we extracted two complex language test datasets, Sophisticated Linguistic Challenge Dataset 1 (SLCD1) and 2 (SLCD2), from HumanML3D, with SLCD1 being more demanding. This approach helps assess our model's ability to handle increasingly intricate linguistic contexts.

\begin{figure}[t]
    \centering
    \includegraphics[width=1\columnwidth]{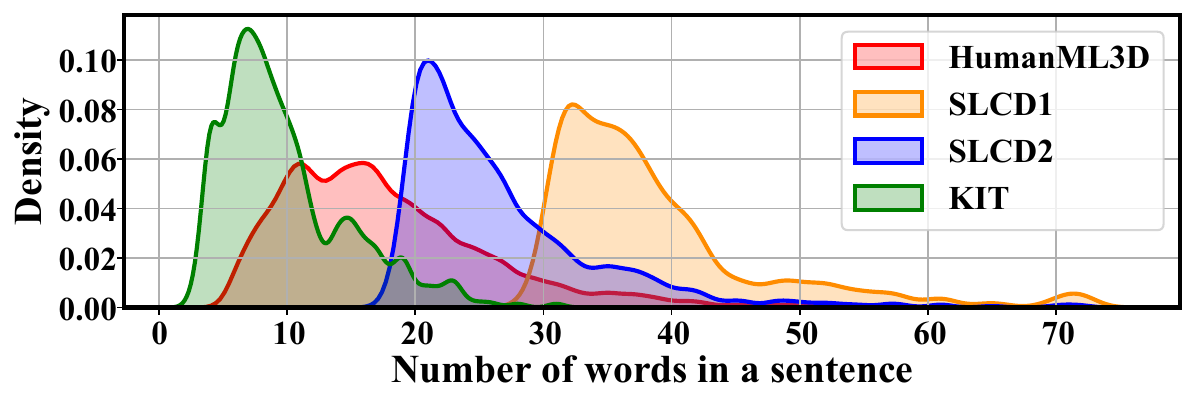}
    \caption{Distribution of text length in the test sets of HumanML3D, SLCD1, SLCD2 and KIT.}
    \label{sentence_length_distribution}
\end{figure}

\subsubsection{Quantitative Results:}
We conducted a comparative study between AMD, and five pre-existing models:MDM, TM2T\cite{guo2022tm2t}, MotionDiffuse, \cite{guo2022generating} and T2M-GPT. First, we evaluated the benchmark model on SLCD1 and SLCD2, respectively.In Table \ref{tab:performance_SLCD1_2}, our model shows outstanding performance in key metrics. Our model outpaces the runner-up, MDM, by 0.712 on the Frechet Inception Distance (FID), and achieves a near-optimal result in Diversity.This highlights our model's superiority in generating quality, realistic motion sequences, and its close performance to the current state-of-the-art in other metrics.
We further tested our model's capabilities on the SLCD2, Human3D, and KIT-ML datasets.
These test sets show a decrease in textual length(assume text length correlates with action complexity.), as seen in Figure \ref{sentence_length_distribution}.
Analysis of Table \ref{tab:performance_SLCD1_2}and \ref{combined-dataset} shows that comparison previous works degrade with complex texts, while our model remains stable, even improving slightly. However, performance on the KIT dataset was below expectations, possibly due to its emphasis on short text descriptions in our selection of complex text datasets.

\begin{figure}[t]
    \centering
    \includegraphics[width=0.75\columnwidth]{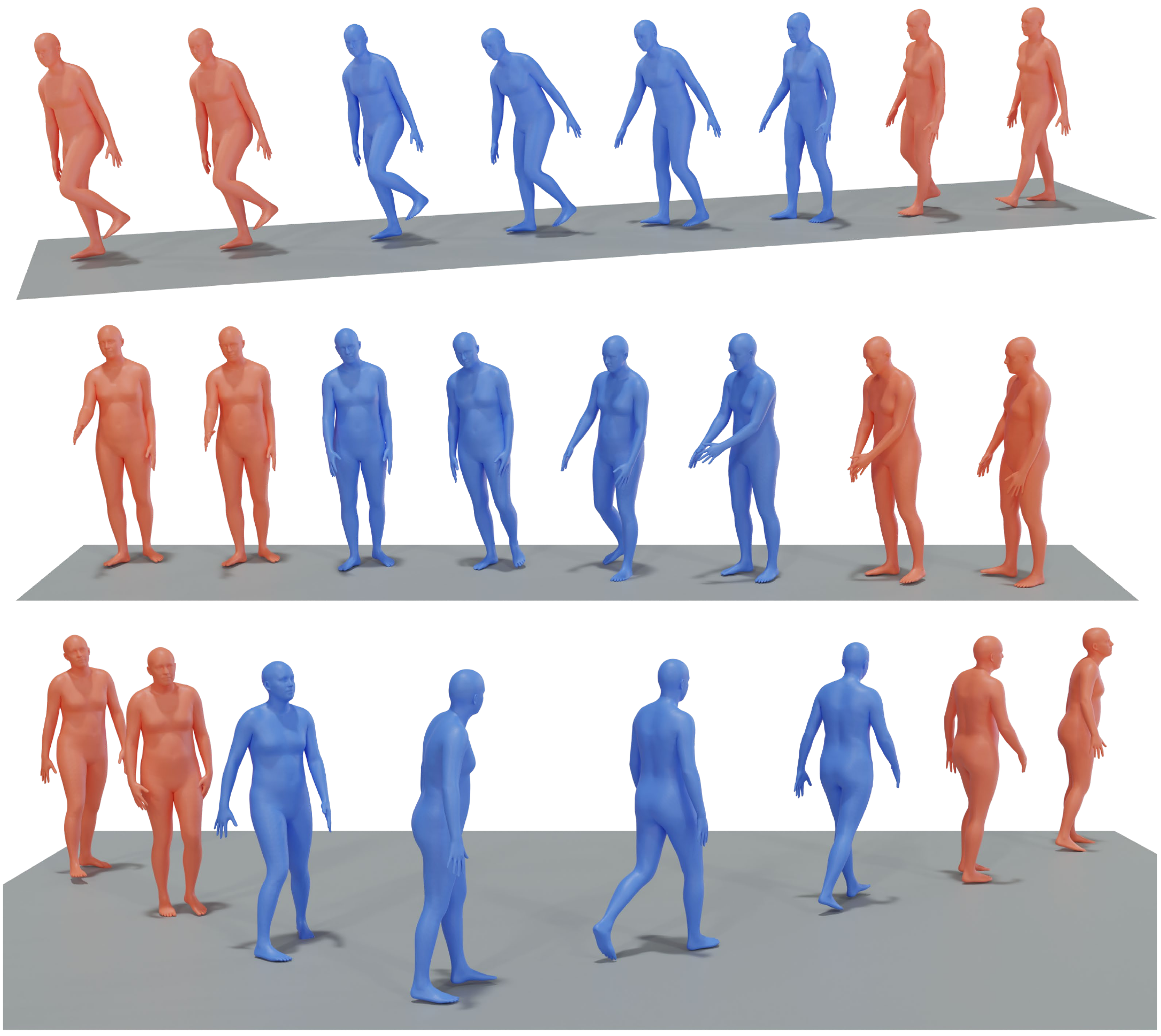}
    \caption{In-between. The blue  is the generate frames and the red is the input frames}
    \label{Motion in-between}
\end{figure}

\begin{figure}[ht]
    \centering
    \includegraphics[width=0.75\columnwidth]{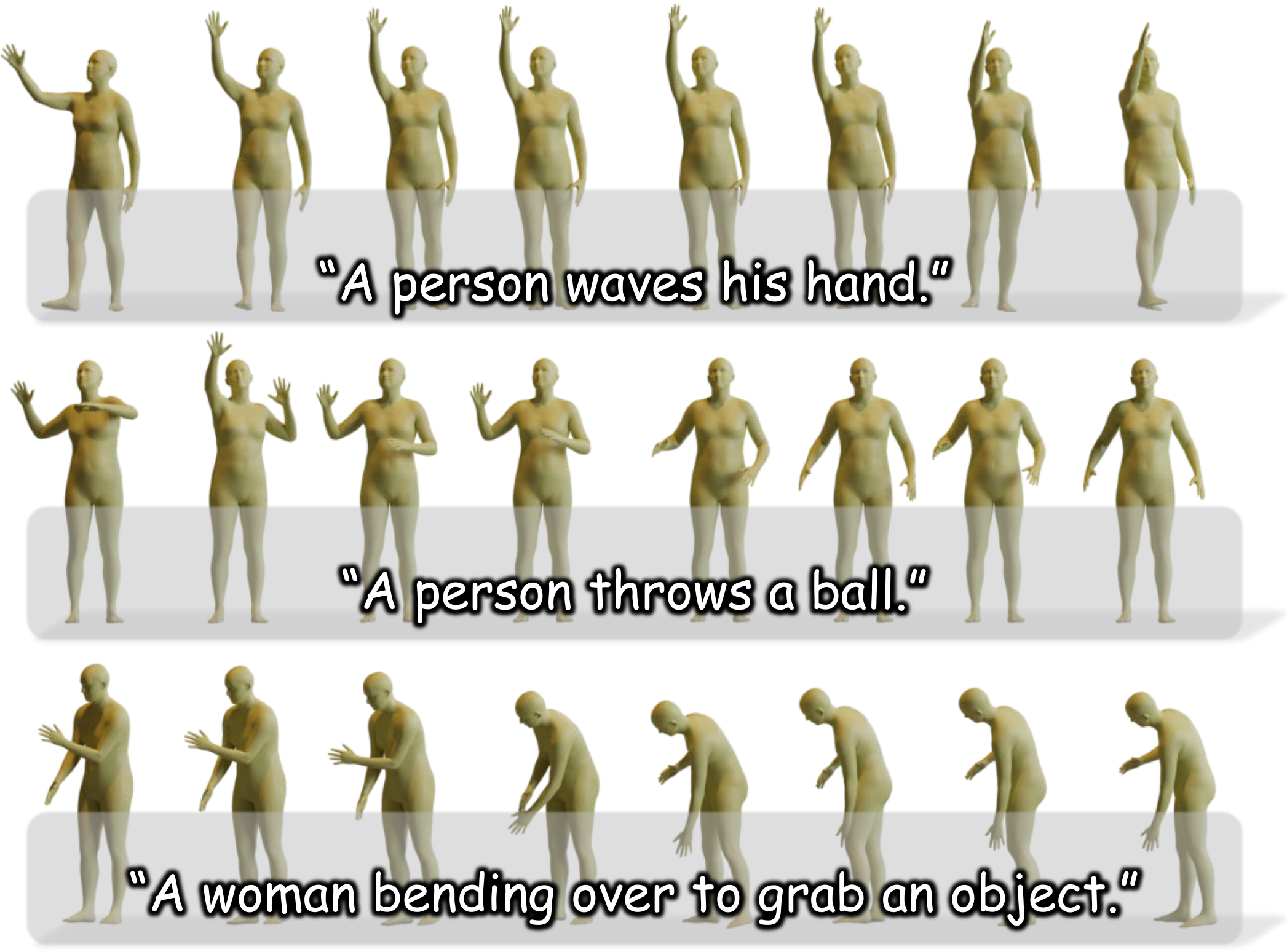}
    \caption{Motion editing. The lower limbs are fixed and only the upper limbs are edited}
    \label{Motion edit}
\end{figure}

\subsubsection{Qualitative Results:}
As shown in Figure \ref{Text to action synthesis_dribbles a ball}, in contrast, the comparison models can only synthesize parts of the text description or generate actions unrelated to the text description. Meanwhile, our AMD can translate text into consistent actions, showing higher text fidelity.

\begin{figure}[ht]
    \centering
    \includegraphics[width=1\columnwidth]{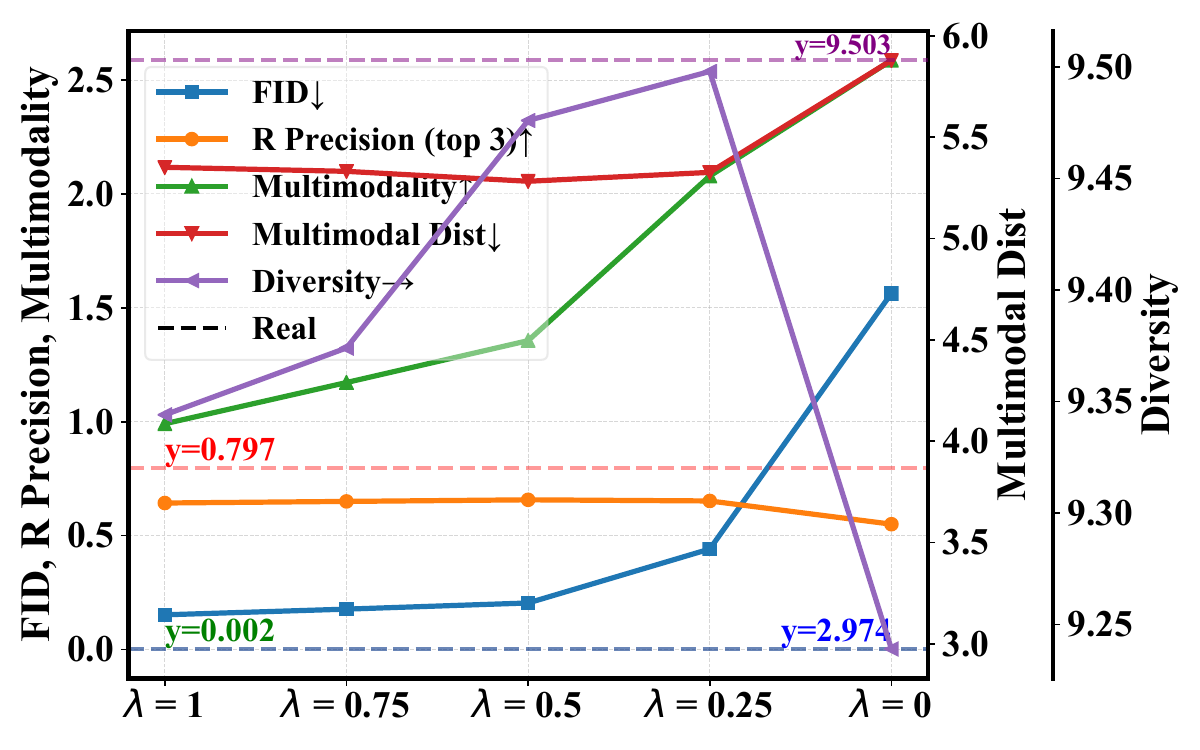}
    \caption{As the $\lambda$ value decreases, changes in the indicators}
    \label{ablation_real}
\end{figure}

\subsection{Ablation study}

We conducted an ablation study on our model, examining no textual decomposition (NTD) and no reference action information (NTDS), with varying values of $\lambda$. Table \ref{tab:abilition} shows that compared to $\lambda=0.5$, all metrics deteriorate to various extents when NTD and NTDS are applied, except for Multimodality. In the comparison between $\lambda=0.5$ and NTD, text decomposition enhances model performance in FID and Multimodal Dist, with moderate improvements in R Precision and Diversity. However, the breakdown of sentences like ``The person moving both hands in an up and down motion'' into discrete steps (1. Raise the left hand. 2. Lower the left hand. 3. Raise the right hand. 4. Lower the right hand.) intensifies constraints on each action generation stage, favoring unimodal distributions and consequently decreasing Multimodality. 

Figure \ref{ablation_real} illustrates the trade-off between two pipelines as \(\lambda\) transitions from 0 to 1. As \(\lambda\) approaches 0, the model exhibits greater randomness, resulting in more diverse outputs that deviate from the dataset distribution, and at \(\lambda = 0\), the absence of guidance from reference actions and the sole operation of pipeline 1 lead to a decrease in the diversity of actions generated. Conversely, as \(\lambda\) approaches 1, the influence of reference action information causes the generated distribution to lean towards reference actions, leading to a notable decrease in the Multimodality metric, and  causes the FID value to drop rapidly at first, then decrease more gradually.This phenomenon shows that \(\lambda\) can adjust the model's fidelity and the diversity of synthesized actions.

\begin{table}[t]
\centering

\setlength\tabcolsep{1pt}
\scalebox{0.9}{
\begin{tabular}{lccccc}
\hline
Method & \begin{tabular}[c]{@{}c@{}}R-Precision\\ (Top 3) $\uparrow$\end{tabular} & FID $\downarrow$ & MM Dist $\downarrow$ & Diversity $\to$ & \begin{tabular}[c]{@{}c@{}}Multi-\\ Modality $\uparrow$\end{tabular} \\ \hline
Real & $0.80^{\pm.00}$ & $0.00^{\pm.00}$ & $2.97^{\pm.01}$ & $9.50^{\pm.07}$ & - \\ \hline
NTD & $0.61^{\pm.01}$ & $0.65^{\pm.07}$ & $5.71^{\pm.03}$ & $9.12^{\pm.08}$ & $1.81^{\pm.11}$ \\
NTDS & $0.54^{\pm.01}$ & $1.39^{\pm.08}$ & $5.98^{\pm.03}$ & $9.08^{\pm.08}$ & \textcolor{red}{\textbf{$2.88^{\pm.10}$}} \\ \hline
Ours & \textcolor{red}{\textbf{$0.66^{\pm.01}$}} & \textcolor{red}{\textbf{$0.20^{\pm.03}$}} & \textcolor{red}{\textbf{$5.28^{\pm.03}$}} & \textcolor{red}{\textbf{$9.48^{\pm.08}$}} & $1.36^{\pm.09}$\\ \hline
\end{tabular}}
\caption{Ablation study on components of AMD.}
\label{tab:abilition}
\end{table}

\subsection{Application on Other Motion Tasks}
In this section, like MDM, we define 25\% of the action's start and end as stable regions, focusing on constructing the middle 50\%. We fixed unedited joints, allowing the model to adjust the rest, particularly the upper body joints. As shown in Figures \ref{Motion in-between} and \ref{Motion edit}, the model can generate smooth, coordinated movements for both intermediate shaping and specific adjustments, meeting set preconditions.

\section{ Conclusion}
In this work, we propose a novel text-to-human-motion generation framework that leverages a fine-tuned ChatGPT-3.5 model to decompose an intricate text into a set of anatomical text. Our approach bypasses the complexity of language descriptions by leveraging the power of LLMs and outperforms the current state-of-the-art models.

\noindent \textbf{limitation} The performance of our model is largely constrained by the broadness of the motion database, as we need to search through the database for reference motion. Though the database only needs to contain the majority of simple actions, but a small database would generally provide low-quality references and affect the performance. In future work, we plan to tackle this issue by distilling simple actions from long video sequences which eases the difficulty of data collection.

\section{Acknowledgments}
This work is supported by the National Natural Science Foundation of China (NSFC No. 62272184). The
computation is completed in the HPC Platform of Huazhong
University of Science and Technology.

\bigskip
\FloatBarrier
\bibliography{aaai24}

\begin{thebibliography}{36}
\providecommand{\natexlab}[1]{#1}

\bibitem[{Ahuja and Morency(2019)}]{ahuja2019language2pose}
Ahuja, C.; and Morency, L.-P. 2019.
\newblock Language2pose: Natural language grounded pose forecasting.
\newblock In \emph{2019 International Conference on 3D Vision (3DV)}, 719--728. IEEE.

\bibitem[{Anil et~al.(2023)Anil, Dai, Firat, Johnson, Lepikhin, Passos, Shakeri, Taropa, Bailey, Chen et~al.}]{anil2023palm}
Anil, R.; Dai, A.~M.; Firat, O.; Johnson, M.; Lepikhin, D.; Passos, A.; Shakeri, S.; Taropa, E.; Bailey, P.; Chen, Z.; et~al. 2023.
\newblock Palm 2 technical report.
\newblock \emph{arXiv preprint arXiv:2305.10403}.

\bibitem[{Ao, Zhang, and Liu(2023)}]{ao2023gesturediffuclip}
Ao, T.; Zhang, Z.; and Liu, L. 2023.
\newblock GestureDiffuCLIP: Gesture diffusion model with CLIP latents.
\newblock \emph{arXiv preprint arXiv:2303.14613}.

\bibitem[{Bhattacharya et~al.(2021)Bhattacharya, Rewkowski, Banerjee, Guhan, Bera, and Manocha}]{bhattacharya2021text2gestures}
Bhattacharya, U.; Rewkowski, N.; Banerjee, A.; Guhan, P.; Bera, A.; and Manocha, D. 2021.
\newblock Text2gestures: A transformer-based network for generating emotive body gestures for virtual agents.
\newblock In \emph{2021 IEEE virtual reality and 3D user interfaces (VR)}, 1--10. IEEE.

\bibitem[{Chen et~al.(2023)Chen, Jiang, Liu, Huang, Fu, Chen, and Yu}]{chen2023executing}
Chen, X.; Jiang, B.; Liu, W.; Huang, Z.; Fu, B.; Chen, T.; and Yu, G. 2023.
\newblock Executing your Commands via Motion Diffusion in Latent Space.
\newblock In \emph{Proceedings of the IEEE/CVF Conference on Computer Vision and Pattern Recognition}, 18000--18010.

\bibitem[{Dabral et~al.(2023)Dabral, Mughal, Golyanik, and Theobalt}]{dabral2023mofusion}
Dabral, R.; Mughal, M.~H.; Golyanik, V.; and Theobalt, C. 2023.
\newblock Mofusion: A framework for denoising-diffusion-based motion synthesis.
\newblock In \emph{Proceedings of the IEEE/CVF Conference on Computer Vision and Pattern Recognition}, 9760--9770.

\bibitem[{Devlin et~al.(2018)Devlin, Chang, Lee, and Toutanova}]{devlin2018bert}
Devlin, J.; Chang, M.-W.; Lee, K.; and Toutanova, K. 2018.
\newblock Bert: Pre-training of deep bidirectional transformers for language understanding.
\newblock \emph{arXiv preprint arXiv:1810.04805}.

\bibitem[{Ghosh et~al.(2021)Ghosh, Cheema, Oguz, Theobalt, and Slusallek}]{ghosh2021synthesis}
Ghosh, A.; Cheema, N.; Oguz, C.; Theobalt, C.; and Slusallek, P. 2021.
\newblock Synthesis of compositional animations from textual descriptions.
\newblock In \emph{Proceedings of the IEEE/CVF international conference on computer vision}, 1396--1406.

\bibitem[{Gong et~al.(2023)Gong, Lian, Chang, Guo, Zuo, Jiang, and Wang}]{gong2023tm2d}
Gong, K.; Lian, D.; Chang, H.; Guo, C.; Zuo, X.; Jiang, Z.; and Wang, X. 2023.
\newblock TM2D: Bimodality Driven 3D Dance Generation via Music-Text Integration.
\newblock \emph{arXiv preprint arXiv:2304.02419}.

\bibitem[{Guo et~al.(2022{\natexlab{a}})Guo, Zou, Zuo, Wang, Ji, Li, and Cheng}]{guo2022generating}
Guo, C.; Zou, S.; Zuo, X.; Wang, S.; Ji, W.; Li, X.; and Cheng, L. 2022{\natexlab{a}}.
\newblock Generating diverse and natural 3d human motions from text.
\newblock In \emph{Proceedings of the IEEE/CVF Conference on Computer Vision and Pattern Recognition}, 5152--5161.

\bibitem[{Guo et~al.(2022{\natexlab{b}})Guo, Zuo, Wang, and Cheng}]{guo2022tm2t}
Guo, C.; Zuo, X.; Wang, S.; and Cheng, L. 2022{\natexlab{b}}.
\newblock Tm2t: Stochastic and tokenized modeling for the reciprocal generation of 3d human motions and texts.
\newblock In \emph{European Conference on Computer Vision}, 580--597. Springer.

\bibitem[{Guo et~al.(2020)Guo, Zuo, Wang, Zou, Sun, Deng, Gong, and Cheng}]{guo2020action2motion}
Guo, C.; Zuo, X.; Wang, S.; Zou, S.; Sun, Q.; Deng, A.; Gong, M.; and Cheng, L. 2020.
\newblock Action2motion: Conditioned generation of 3d human motions.
\newblock In \emph{Proceedings of the 28th ACM International Conference on Multimedia}, 2021--2029.

\bibitem[{Ho, Jain, and Abbeel(2020)}]{ho2020denoising}
Ho, J.; Jain, A.; and Abbeel, P. 2020.
\newblock Denoising diffusion probabilistic models.
\newblock \emph{Advances in neural information processing systems}, 33: 6840--6851.

\bibitem[{Kim, Kim, and Choi(2023)}]{kim2023flame}
Kim, J.; Kim, J.; and Choi, S. 2023.
\newblock Flame: Free-form language-based motion synthesis \& editing.
\newblock In \emph{Proceedings of the AAAI Conference on Artificial Intelligence}, 8255--8263.

\bibitem[{Kingma and Welling(2013)}]{kingma2013auto}
Kingma, D.~P.; and Welling, M. 2013.
\newblock Auto-encoding variational bayes.
\newblock \emph{arXiv preprint arXiv:1312.6114}.

\bibitem[{Lee et~al.(2019)Lee, Yang, Liu, Wang, Lu, Yang, and Kautz}]{lee2019dancing}
Lee, H.-Y.; Yang, X.; Liu, M.-Y.; Wang, T.-C.; Lu, Y.-D.; Yang, M.-H.; and Kautz, J. 2019.
\newblock Dancing to music.
\newblock \emph{Advances in neural information processing systems}, 32.

\bibitem[{Lee, Moon, and Lee(2023)}]{lee2023multiact}
Lee, T.; Moon, G.; and Lee, K.~M. 2023.
\newblock MultiAct: Long-Term 3D Human Motion Generation from Multiple Action Labels.
\newblock In \emph{Proceedings of the AAAI Conference on Artificial Intelligence}, 1231--1239.

\bibitem[{Lin et~al.(2018)Lin, Wu, Corona, Tai, Huang, and Mooney}]{lin2018generating}
Lin, A.~S.; Wu, L.; Corona, R.; Tai, K.; Huang, Q.; and Mooney, R.~J. 2018.
\newblock Generating animated videos of human activities from natural language descriptions.
\newblock \emph{Learning}, 2018(1).

\bibitem[{Lucas et~al.(2022)Lucas, Baradel, Weinzaepfel, and Rogez}]{lucas2022posegpt}
Lucas, T.; Baradel, F.; Weinzaepfel, P.; and Rogez, G. 2022.
\newblock Posegpt: Quantization-based 3d human motion generation and forecasting.
\newblock In \emph{European Conference on Computer Vision}, 417--435. Springer.

\bibitem[{Mahmood et~al.(2019)Mahmood, Ghorbani, Troje, Pons-Moll, and Black}]{mahmood2019amass}
Mahmood, N.; Ghorbani, N.; Troje, N.~F.; Pons-Moll, G.; and Black, M.~J. 2019.
\newblock AMASS: Archive of motion capture as surface shapes.
\newblock In \emph{Proceedings of the IEEE/CVF international conference on computer vision}, 5442--5451.

\bibitem[{Mandery et~al.(2015)Mandery, Terlemez, Do, Vahrenkamp, and Asfour}]{mandery2015kit}
Mandery, C.; Terlemez, {\"O}.; Do, M.; Vahrenkamp, N.; and Asfour, T. 2015.
\newblock The KIT whole-body human motion database.
\newblock In \emph{2015 International Conference on Advanced Robotics (ICAR)}, 329--336. IEEE.

\bibitem[{Mukai and Kuriyama(2005)}]{mukai2005geostatistical}
Mukai, T.; and Kuriyama, S. 2005.
\newblock Geostatistical motion interpolation.
\newblock In \emph{ACM SIGGRAPH 2005 Papers}, 1062--1070. SIGGRAPH.

\bibitem[{Petrovich, Black, and Varol(2022)}]{petrovich2022temos}
Petrovich, M.; Black, M.~J.; and Varol, G. 2022.
\newblock TEMOS: Generating diverse human motions from textual descriptions.
\newblock In \emph{European Conference on Computer Vision}, 480--497. Springer.

\bibitem[{Ren et~al.(2023)Ren, Pan, Zhou, and Kang}]{ren2023diffusion}
Ren, Z.; Pan, Z.; Zhou, X.; and Kang, L. 2023.
\newblock Diffusion motion: Generate text-guided 3d human motion by diffusion model.
\newblock In \emph{ICASSP 2023-2023 IEEE International Conference on Acoustics, Speech and Signal Processing (ICASSP)}, 1--5. IEEE.

\bibitem[{Rombach et~al.(2022)Rombach, Blattmann, Lorenz, Esser, and Ommer}]{rombach2022high}
Rombach, R.; Blattmann, A.; Lorenz, D.; Esser, P.; and Ommer, B. 2022.
\newblock High-resolution image synthesis with latent diffusion models.
\newblock In \emph{Proceedings of the IEEE/CVF conference on computer vision and pattern recognition}, 10684--10695.

\bibitem[{Rose, Cohen, and Bodenheimer(1998)}]{rose1998verbs}
Rose, C.; Cohen, M.~F.; and Bodenheimer, B. 1998.
\newblock Verbs and adverbs: Multidimensional motion interpolation.
\newblock \emph{IEEE Computer Graphics and Applications}, 18(5): 32--40.

\bibitem[{Saharia et~al.(2022)Saharia, Chan, Saxena, Li, Whang, Denton, Ghasemipour, Gontijo~Lopes, Karagol~Ayan, Salimans et~al.}]{saharia2022photorealistic}
Saharia, C.; Chan, W.; Saxena, S.; Li, L.; Whang, J.; Denton, E.~L.; Ghasemipour, K.; Gontijo~Lopes, R.; Karagol~Ayan, B.; Salimans, T.; et~al. 2022.
\newblock Photorealistic text-to-image diffusion models with deep language understanding.
\newblock \emph{Advances in Neural Information Processing Systems}, 35: 36479--36494.

\bibitem[{Shafir et~al.(2023)Shafir, Tevet, Kapon, and Bermano}]{shafir2023human}
Shafir, Y.; Tevet, G.; Kapon, R.; and Bermano, A.~H. 2023.
\newblock Human motion diffusion as a generative prior.
\newblock \emph{arXiv preprint arXiv:2303.01418}.

\bibitem[{Song, Meng, and Ermon(2020)}]{song2020denoising}
Song, J.; Meng, C.; and Ermon, S. 2020.
\newblock Denoising diffusion implicit models.
\newblock \emph{arXiv preprint arXiv:2010.02502}.

\bibitem[{Tevet et~al.(2022)Tevet, Raab, Gordon, Shafir, Cohen-Or, and Bermano}]{tevet2022human}
Tevet, G.; Raab, S.; Gordon, B.; Shafir, Y.; Cohen-Or, D.; and Bermano, A.~H. 2022.
\newblock Human motion diffusion model.
\newblock \emph{arXiv preprint arXiv:2209.14916}.

\bibitem[{Yuan et~al.(2022)Yuan, Song, Iqbal, Vahdat, and Kautz}]{yuan2022physdiff}
Yuan, Y.; Song, J.; Iqbal, U.; Vahdat, A.; and Kautz, J. 2022.
\newblock Physdiff: Physics-guided human motion diffusion model.
\newblock \emph{arXiv preprint arXiv:2212.02500}.

\bibitem[{Zhang et~al.(2023{\natexlab{a}})Zhang, Zhang, Cun, Huang, Zhang, Zhao, Lu, and Shen}]{zhang2023t2m}
Zhang, J.; Zhang, Y.; Cun, X.; Huang, S.; Zhang, Y.; Zhao, H.; Lu, H.; and Shen, X. 2023{\natexlab{a}}.
\newblock T2m-gpt: Generating human motion from textual descriptions with discrete representations.
\newblock \emph{arXiv preprint arXiv:2301.06052}.

\bibitem[{Zhang et~al.(2022)Zhang, Cai, Pan, Hong, Guo, Yang, and Liu}]{zhang2022motiondiffuse}
Zhang, M.; Cai, Z.; Pan, L.; Hong, F.; Guo, X.; Yang, L.; and Liu, Z. 2022.
\newblock Motiondiffuse: Text-driven human motion generation with diffusion model.
\newblock \emph{arXiv preprint arXiv:2208.15001}.

\bibitem[{Zhang et~al.(2023{\natexlab{b}})Zhang, Guo, Pan, Cai, Hong, Li, Yang, and Liu}]{zhang2023remodiffuse}
Zhang, M.; Guo, X.; Pan, L.; Cai, Z.; Hong, F.; Li, H.; Yang, L.; and Liu, Z. 2023{\natexlab{b}}.
\newblock ReMoDiffuse: Retrieval-Augmented Motion Diffusion Model.
\newblock \emph{arXiv preprint arXiv:2304.01116}.

\bibitem[{Zhang and Tang(2022)}]{zhang2022wanderings}
Zhang, Y.; and Tang, S. 2022.
\newblock The wanderings of odysseus in 3d scenes.
\newblock In \emph{Proceedings of the IEEE/CVF Conference on Computer Vision and Pattern Recognition}, 20481--20491.

\bibitem[{Zhao et~al.(2023)Zhao, Liu, Ren, Dai, and Sebe}]{zhao2023modiff}
Zhao, M.; Liu, M.; Ren, B.; Dai, S.; and Sebe, N. 2023.
\newblock Modiff: Action-conditioned 3d motion generation with denoising diffusion probabilistic models.
\newblock \emph{arXiv preprint arXiv:2301.03949}.

\end{thebibliography}

\end{document}